\documentclass{bjmc}   
\usepackage{natbib}
\usepackage[T1]{fontenc}
\usepackage[main=english,ukrainian]{babel}

\usepackage{url}	    
\usepackage{todonotes}
 \usepackage{multirow}
\usepackage{booktabs}	   	

\begin{document}

%
\title{From Zero to Production: Baltic-Ukrainian Machine Translation Systems to Aid Refugees}
\titlerunning{Baltic-Ukrainian MT Systems to Aid Refugees}  

\author{Toms BERGMANIS and Mārcis PINNIS}
\authorrunning{Bergmanis and Pinnis}  
\institute{Tilde, Vienības gatve 75A, Riga, Latvia, LV-1004 
}

\emails{\{name.surname\}@tilde.lv}
\maketitle
\begin{abstract} 
In this paper, we examine the development and usage of six low-resource machine translation systems translating between the Ukrainian language and each of the official languages of the Baltic states. We developed these systems in reaction to the escalating Ukrainian refugee crisis caused by the Russian military aggression in Ukraine in the hope that they might be helpful for refugees and public administrations. 
Now, two months after MT systems were made public, we analyze their usage patterns and statistics. Our findings show that the Latvian-Ukrainian and Lithuanian-Ukrainian systems are integrated into the public services of Baltic states, leading to more than 127 million translated sentences for the Lithuanian-Ukrainian system. Motivated by these findings, we further enhance our MT systems by better Ukrainian toponym translation and publish an improved version of the Lithuanian-Ukrainian system. 
\end{abstract}

\section{Introduction} 
On February 20, 2014, Russian Federation started military aggression against Ukraine \citep{cosgrove2020russian}. Eight years later, on February 24, 2022, following a Russian military build-up on the Russia–Ukraine border, Russian aggression culminated in a full-scale invasion of Ukraine.\footnote{Accessed May 16, 2022 https://news.un.org/en/focus/ukraine} As of May 2022, more than 6.1 million refugees have fled Ukraine.\footnote{Accessed May 16, 2022 \url{https://data2.unhcr.org/en/situations/ukraine}} The majority of refugees have left Ukraine for one of the seven neighboring countries. Still, many seek shelter in other countries, including the Baltic states.\footnote{Accessed May 5, 2022  \url{https://en.wikipedia.org/wiki/2022_Ukrainian_refugee_crisis#Other_European_countries}} The influx of Ukrainian refugees poses a new challenge for communication between individuals and governmental bodies in the Baltic states.

In this paper, we examine six low-resource machine translation (MT) systems translating between the Ukrainian language and each of the official languages of the Baltic states. Their development took place in the wake of the escalating Ukrainian refugee crisis shortly after the Russian invasion of Ukraine. Thus it was motivated by apprehension for the future rather than a clear vision of how they might be used. Now, after MT systems have been online for more than two months, we analyze their usage statistics and draw conclusions for what are the aspects of MT integration in the public services, which have led to more than 127 million translated sentences for the Lithuanian-Ukrainian system, while the Latvian-Ukrainian system has been used seemingly relatively little having translated only 138 thousand sentences. 


\section{Machine Translation Systems} \label{section:systems}
Due to data scarcity for the language pairs involving Ukrainian and the languages of the Baltic states, we use two data augmentation methods -- one that enables dynamic terminology integration and another that allows training MT models that are more robust to unknown tokens and rare words. For terminology integration, we prepare data with Target Lemma Annotations (TLA) \citep{bergmanis-pinnis-2021-facilitating}, while for the robustness, we use synthetic data augmentation as proposed by \cite{pinnis2017neural}.

For system training, we use the \textit{Marian} neural machine translation (NMT) toolkit by \citep{mariannmt}. We train standard NMT systems that largely follow the Transformer \citep{vaswani2017attention} \textit{base} model configuration. The only departures from the standard configuration are the changes necessary for TLA support during training and inference. For the \textit{Marian} toolkit, they were described in \cite{bergmanis-pinnis-2021-dynamic}. Specifically, we employ the source-side factors using factor embeddings of dimensionality of 8 and concatenate them with subword embeddings. We also increase the delay of updates for the optimiser\footnote{The delay of updates for the optimiser can be enabled in Marian using the \textit{--optimizer-delay} parameter.} (from 16 to 24 batches) and set the maximum sequence length to 196 tokens. The increased sequence length accounts for longer input sequences caused by the additional TLA tokens. On the other hand, the increased optimizer delay negates the effect of the smaller effective batch size due to fewer sentences fitting in the workspace memory-based batch because of their increased maximum length. Furthermore, all models are trained using the guided alignment functionality of the \textit{Marian} toolkit.

\begin{table}[!t]
\caption{Parallel data statistics for each language pair before and after filtering as well as after synthetic data generation.}
\label{tab:data-stats}
\centering
\begin{tabular}{lcccc}
\toprule
\multicolumn{1}{c}{\begin{tabular}[c]{@{}c@{}}\end{tabular}} &
  \multicolumn{1}{c}{\begin{tabular}[c]{@{}c@{}}\textbf{Before}\\ \textbf{filtering}\end{tabular}} &
  \multicolumn{1}{c}{\begin{tabular}[c]{@{}c@{}}\textbf{After}\\ \textbf{filtering}\end{tabular}} &
  \multicolumn{1}{c}{\begin{tabular}[c]{@{}c@{}}\textbf{After unknown}\\ \textbf{data generation}\end{tabular}} &
  \multicolumn{1}{c}{\begin{tabular}[c]{@{}c@{}}\textbf{After} \\ \textbf{TLA}\end{tabular}} \\ \midrule
UK-LV & 4.0M  & 2.3M  & 4.5M  & 8.9M  \\
UK-ET & 5.1M  & 4.0M  & 7.8M  & 15.5M \\
UK-LT & 5.9M  & 4.7M  & 9.3M  & 18.6M \\\bottomrule
\end{tabular}
\end{table}

To train MT systems, we mostly use publicly available parallel data from the Tatoeba Challenge \citep{tiedemann2020tatoeba} corpus. This constitutes 69\%, 70\%, and 74\% of all parallel data for Latvian-Ukrainian, Estonian-Ukrainian, and Lithuanian-Ukrainian respectively. The remaining data were acquired from proprietary data sources. Data statistics are depicted in Table~\ref{tab:data-stats}. We filtered all parallel data using parallel data filtering methods by \cite{pinnis-2018-tildes} and then performed pre-processing, which included the following steps: normalisation of whitespaces and punctuation (e.g., quotation marks, apostrophes, hyphens, etc.), identification of non-translatable entities (e.g., e-mails, file paths, complex identifiers are replaced with placeholders), tokenisation, truecasing, synthetic unknown data generation \citep{pinnis2017neural}, byte-pair encoding \citep{sennrich2015neural}, and finally TLA.

For validation of our MT systems during training and for evaluation, we use the \textit{dev} and \textit{devtest} splits of FLORES-101---an evaluation benchmark specially created for low-resource language pairs \citep{goyal2022flores}. We use the standard splits, which consist of 997 and 1012 sentences large validations and evaluation sets respectively that are parallel across all four languages.

\begin{table}[]
\centering
\caption{Results of automatic evaluation using SacreBLEU implementation of ChrF, BLEU and TER metrics. }
\begin{tabular}{llrrrrllrrr}
\toprule
                       &               & \textbf{ChrF} $\uparrow$ & \textbf{BLEU} $\uparrow$ & \textbf{TER} $\downarrow$  &  &                        &               & \textbf{ChrF} $\uparrow$ & \textbf{BLEU} $\uparrow$ & \textbf{TER} $\downarrow$  \\ \midrule
\multirow{3}{*}{UK-LV} & ~eTranslation & 53.7 & 23.0 & 65.4 &  & \multirow{3}{*}{LV-UK} & ~eTranslation & 51.4 & 21.2 & 69.8 \\
                       & ~Google       & 57.0 & 26.9 & 60.7 &  &                        & ~Google       & 53.0 & 22.6 & 66.0 \\
                       & ~This Work    & 52.2 & 21.2 & 67.3 &  &                        & ~This Work    & 47.6 & 17.5 & 73.0 \\  \midrule
\multirow{3}{*}{UK-LT} & ~eTranslation & 54.0 & 21.4 & 69.0 &  & \multirow{3}{*}{LT-UK} & ~eTranslation & 49.1 & 19.1 & 70.7 \\
                       & Google        & 56.6 & 24.4 & 64.5 &  &                        & ~Google       & 50.8 & 21.0 & 67.9 \\
                       & This Work     & 54.5 & 21.7 & 67.7 &  &                        & ~This Work    & 49.4 & 18.8 & 71.0 \\  \midrule
\multirow{3}{*}{UK-ET} & ~eTranslation & 53.6 & 19.4 & 69.2 &  & \multirow{3}{*}{ET-UK} & ~eTranslation & 50.6 & 20.8 & 68.6 \\
                       & Google        & 56.2 & 22.0 & 65.7 &  &                        & ~Google       & 52.4 & 23.0 & 65.4 \\
                       & This Work     & 53.5 & 19.4 & 69.2 &  &                        & ~This Work    & 49.5 & 19.5 & 70.5 \\
\bottomrule 
\end{tabular}
\label{table:automaticResults}
\end{table}

\subsection{Automatic Evaluation}
We compare our systems with Google Translate\footnote{Accessed May 16, 2022 \url{https://translate.google.com}} and eTranslation\footnote{Accessed May 16, 2022 \url{https://ec.europa.eu/info/resources-partners/machine-translation-public-administrations-etranslation_en}}. We compare against Google Translate because, for many people, it is the go-to MT service provider when the amount of text to be translated is small. However, Google Translate is not free of charge when translation volumes exceed a certain limit. Thus we also compare against eTranslation -- the MT service provider of the European Commission. eTranslation is free of charge for European small and medium-sized enterprises, employees of public administrations across the European Union and public sector service providers. 

Table~\ref{table:automaticResults} shows results of automatic evaluation using SacreBLEU \citep{post2018call} implementation of three MT evaluation metrics--- ChrF\footnote{SacreBLEU hash: nrefs:1\textbar case:mixed\textbar eff:yes \textbar nc:6\textbar nw:0\textbar space:no\textbar version:2.0.0} \citep{popovic2015chrf}, BLEU\footnote{SacreBLEU hash: nrefs:1\textbar case:mixed\textbar eff:no \textbar tok:13a \textbar smooth:exp \textbar version:2.0.0} \citep{papineni2002bleu}, and TER\footnote{SacreBLEU hash:   nrefs:1\textbar case:lc\textbar tok:tercom\textbar norm:no \textbar punct:yes \textbar asian:no \textbar version:2.0.0} \citep{snover2006study}.

The automatic evaluation using ChrF, which is the most suitable metric for morphologically complex languages \citep{kocmi-etal-2021-ship} such as the languages considered in this work, shows that Google Translate performs the best. Our systems compare to eTranslation in the range from marginally better for Ukrainian-Lithuanian and Lithuanian-Ukrainian directions to substantially worse for Ukrainian-Latvian. While these results do not favor our MT systems, they serve as a sanity check. Even though our systems are a one-shot attempt at developing MT systems for a set of low-resource language pairs, they are, to an extent, comparable to other publicly available alternatives.

\begin{figure}[ht!]
\centering
\includegraphics[width=1.0\textwidth]{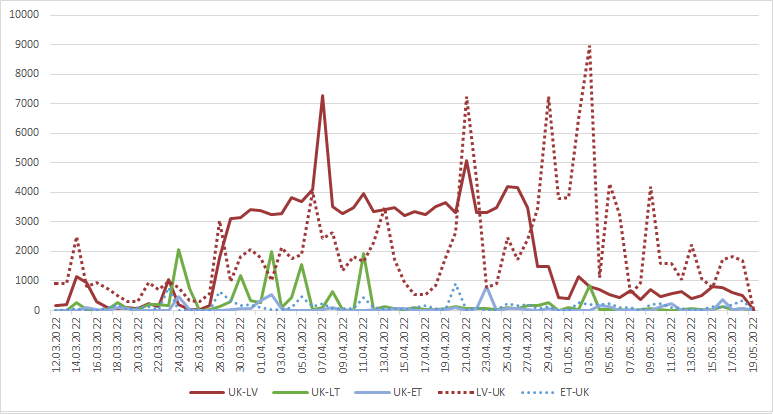}
\caption{The number of sentences translated daily for MT systems translating between Ukrainian and languages of Baltic states, except for the Lithuanian-Ukrainian system for which data is plotted separately (see Figure~\ref{fig:lt_uk}). Statistics are from March 11, 2022, to May 19, 2022.}
\label{fig:uk_baltic}
\end{figure}
\begin{figure}[ht!]
\centering
\includegraphics[width=1.0\textwidth]{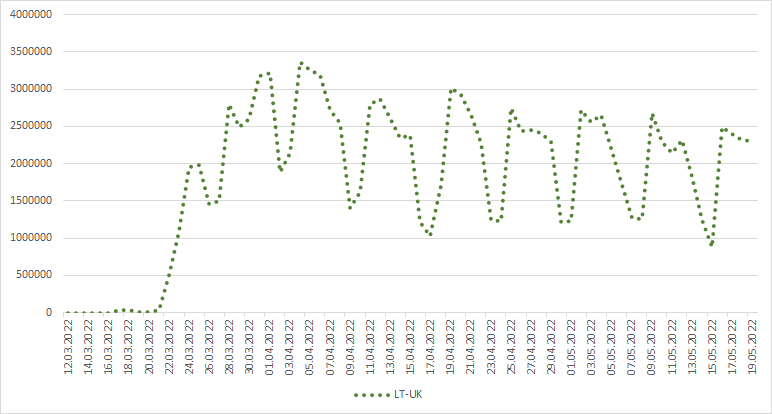}
\caption{The number of sentences translated daily for Lithuanian-Ukrainian MT system. Statistics are from March 11, 2022, to May 19, 2022.}
\label{fig:lt_uk}
\end{figure}

\section{Usage of MT Systems}\label{section:usage}
We published our MT systems on March 11, 2022, which means that they have been online for more than two months at the time of writing the paper. In this section, we aim to analyze how our systems are used and who are their users. Figure~\ref{fig:uk_baltic} shows the number of translated sentences by each system. Due to its large translation volume, usage statistics for the Lithuanian-Ukrainian MT system are plotted separately in Figure~\ref{fig:lt_uk}. The graphs show that Estonian systems were used the least, having translated only about five thousand sentences from Ukrainian to Estonian and almost twice as much from Estonian to Ukrainian. Slightly higher usage numbers are evident for Latvian systems, which have processed more than 138 thousand and 132 thousand Latvian-Ukrainian and Ukrainian-Latvian translation sentences, respectively. Although the Ukrainian-Lithuanian system has translated only about 16 thousand sentences, the Lithuanian-Ukrainian system has had the highest demand as it has translated more than 127 million sentences.

\begin{figure}[ht!]
\centering
\includegraphics[width=1.0\textwidth]{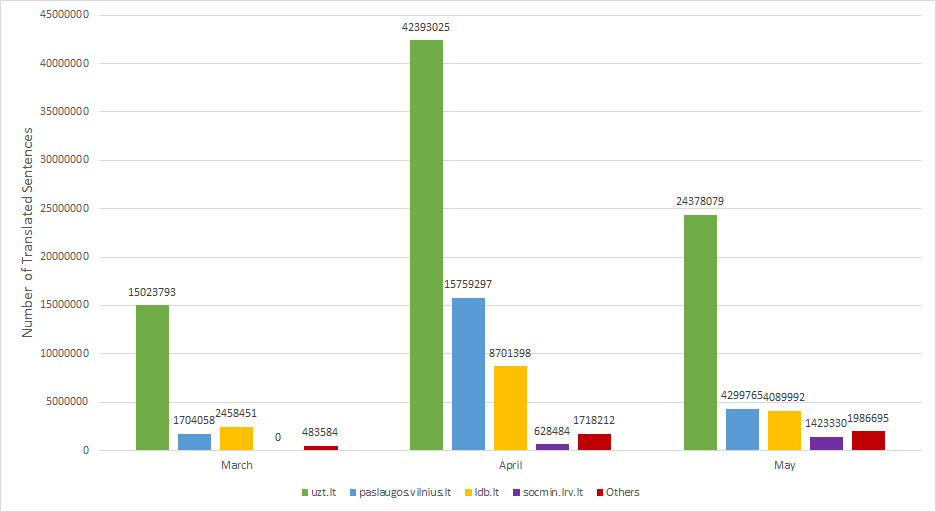}
\caption{The number of sentences translated by the top users of the Lithuanian-Ukrainian MT system. Statistics are from March 11, 2022, to May 19, 2022.}
\label{fig:lt_domains}
\end{figure}

Analyzing through what channels our systems are accessed reveals that the Latvian systems are only one-quarter of the time used by our paid clients. However, they are most often used via the Latvian language technology platform \texttt{\url{hugo.lv}}, which is popular among freelance translators and governmental organizations.
As for Lithuanian systems, the users translating from Ukrainian into Lithuanian have almost exclusively used our public translation platform \texttt{translate.tilde.com}, which allows for speculation that individual users made these translation requests, most likely translating text snippets from news and social media. The system for the opposite translation direction is translating from Lithuanian into Ukrainian and has almost entirely been used via \texttt{Tilde Web Translation Widget}. To understand the 127 million sentences large volume of translated sentences, we inspect the distribution of translated sentences by their source website (see Figure~\ref{fig:lt_domains}). All websites using this MT system are related to the Lithuanian government. The top websites are \texttt{\url{uzt.lt}} and \texttt{\url{ldb.lt}}, which are services of the employment agency of Lithuania, \texttt{\url{paslaugos.vilnius.lt}}, which is the Vilnius City Council services' page and \texttt{\url{socmin.lrv.lt}}, which is the Ministry of Social Security and Labor of the Republic of Lithuania. This analysis reveals that, at least as far as the usage of the Lithuanian-Ukrainian MT system is concerned, even if just a little, our work has helped the people in need to access help and social services.

It is also important to note that the difference of the usage levels for Latvian and Lithuanian systems can be explained with how the systems are used in Latvia and Lithuania. In Lithuania, the Lithuanian-Ukrainian system is (mostly) used to translate governmental websites. Whenever a user (a citizen, a refugee, or a tourist) accesses a certain page in a website, its content is translated by the MT system. This generates high numbers of translation requests. However, this method allows to provide instant multilinguality in a website regardless of which page a user wants to see. In Latvia, the systems are mostly used by translators and public service officials in post-editing scenarios. This means that different from Lithuania where we can grasp a rough estimate of how many end-users consume the translations, in Latvia we only know how many unique sentences were translated to create content in a different language. We cannot estimate how many end-users might have consumed that content. However, the volume is still substantial for post-editing scenarios.

\begin{table}[]
\centering
\caption{Results of automatic evaluation using the SacreBLEU implementation of ChrF, BLEU and TER metrics for Lithuanian-Ukrainian MT systems. \textit{This Work's BT System} refers to the system developed using back-tanslated data, while \textit{This Work's Baseline} refers to MT systems described in Section~\ref{section:systems}. \textbf{*}~denotes that the result difference between this and the system trained on back-translated data is statistically significant according to SacreBLEU's paired bootstrap re-sampling test.}
\label{table:results2}
\begin{tabular}{llll}
\toprule
                     & \textbf{ChrF} $\uparrow$ & \textbf{BLEU} $\uparrow$ & \textbf{TER} $\downarrow$  \\
                     \midrule
eTranslation         & 49.1* & 19.1 & 70.7 \\
Google Translate     & 50.8* & 21.0 & 67.9 \\
This Work's Baseline & 49.4* & 18.8 & 71.0 \\
This Work's BT System  & 50.0 & 19.3 & 70.7 \\
\midrule \\
\end{tabular}

\end{table}
\section{Step Forward}
\subsection{Lithuanian-Ukrainian MT System with Back-translated Data}
In Section~\ref{section:usage}, we established that the Lithuanian-Ukrainian MT system is used the most as it has translated 127 million sentences helping Ukrainians in Lithuania to find jobs and access social services. 
Besides, unlike the Latvian-Ukrainian system, which is primarily used in post-editing scenarios, the translations of the Lithuanian-Ukrainian system reach its users without the supervision of professional translators.
Therefore, we aim to deliver better technology where the people use it the most and retrain our Lithuanian-Ukrainian MT system.

\subsubsection{Machine Translation Systems}
We use the Ukrainian-Lithuanian MT system to create synthetic parallel data by back-translation \citep{sennrich-etal-2016-improving} of monolingual Ukrainian data. For data sources, we use the 2008 to 2021 News Crawl\footnote{\url{https://data.statmt.org/news-crawl/uk}} corpus provided by the Machine Translation Group at the University of Edinburgh and the RSS News, Newscrawl, and Wikipedia corpora\footnote{\url{https://wortschatz.uni-leipzig.de/en/download/Ukrainian}} collected by the University of Leipzig \citep{goldhahn-etal-2012-building}. We also use the Ukrainian side of the Ukrainian-English Wikimedia \citep{tiedemann-2012-parallel}, TED 2020 \citep{reimers-2020-multilingual-sentence-bert}, and OpenSubtitles v2018 \citep{lison-tiedemann-2016-opensubtitles2016} corpora from Opus\footnote{\url{https://opus.nlpl.eu/}} \citep{tiedemann-2012-parallel}. Altogether these corpora amount to around 11.4 million sentences.

As before, we continue by using the synthetic data augmentation \citep{pinnis2017neural}, which nearly doubles the number of sentences to about 21 million. 
We then translate this data into Lithuanian and use parallel data filters by \cite{pinnis-2018-tildes} to get rid of noisy and poor quality translations, which leaves us with around 19 million back-translated sentences. Finally, we add this data to the data we used to train the baseline system (see Section~\ref{section:systems}) to obtain a total of about 37.8 million sentences. We then use the same configuration as described in Section~\ref{section:systems} with an exception that we increase optimizer delay from 24 to 64.

\subsubsection{Automatic Evaluation}
Table~\ref{table:results2} shows a comparison of the baseline MT systems from Section~\label{section:systems} and the newly created Lithuanian-Ukrainian MT system. The new system achieves the second best results, conceding only to Google Translate, which is still 0.8 ChrF points better. However, the results also show that using back-translated data helps to yield statistically significant improvements in translation quality over the other two baselines.

\begin{table}[ht!]
\centering
 \caption{Examples of Ukrainian toponyms and their translations previously represented via their pronunciation in Russian.}
 \label{table:toponym_translations}
\small
\begin{tabular}{lllllll}
\toprule
\textbf{UK}         & \multicolumn{2}{c}{\textbf{EN}}        & \multicolumn{2}{c}{\textbf{LV}}     & \multicolumn{2}{c}{\textbf{LT}}       \\ 
           & \textbf{New}     & \textbf{Obsolete}  & \textbf{New}   & \textbf{Obsolete}& \textbf{New}     & \textbf{Obsolete} \\\midrule
\foreignlanguage{ukrainian}{Київ}       & Kyiv         & Kiev           & Kijiva     & Kijeva        & Kyjivas      & Kyjevas       \\
\foreignlanguage{ukrainian}{Харків}     & Kharkiv      & Kharkov        & Harkiva    & Harkova       & Charkivas    & Charkovas     \\
\foreignlanguage{ukrainian}{Одеса}      & Odessa       & Odesa          & Odesa      & -             & Odesa        & -             \\
\foreignlanguage{ukrainian}{Дніпро}     & Dnipro       & Dnipropetrovsk & Dnipro     & Dņepro        & Dnipras      & Dniepras      \\
\foreignlanguage{ukrainian}{Донецьк}    & Donetsk      & -              & Donecka    & Doņecka       & Doneckas     & -             \\
\foreignlanguage{ukrainian}{Запоріжжя}  & Zaporizhzhia & -              & Zaporižja  & Zaporožje     & Zaporižė     & Zaparožė      \\
\foreignlanguage{ukrainian}{Львів}      & Lviv         & Lvov           & Ļviva      & Ļvova         & Lvivas       & Lvovas        \\
\foreignlanguage{ukrainian}{Кривий Ріг} & Kryvyi Rih   & -              & Krivijriha & Krivojroga    & Kryvyj Rihas & -             \\
\foreignlanguage{ukrainian}{Миколаїв}   & Mykolaiv     & Nikolaev       & Mikolajiva & Mikolajeva    & Mykolajevas  & -             \\
\foreignlanguage{ukrainian}{Луганськ}   & Luhansk      & Lugansk        & Luhanska   & Luganska      & Luhanskas    & -             \\
\foreignlanguage{ukrainian}{Макіївка}   & Makiivka     & Makiyivka      & Makijivka  & -             & Makijivka    & -             \\
\foreignlanguage{ukrainian}{Вінниця}   & Vinnytsia    & -              & Vinnica    & Viņņica       & Vinica       & -             \\
\foreignlanguage{ukrainian}{Чернігів}   & Chernihiv    & Chernigov      & Černihiva  & Čerņigova     & Černyhivas   & Černyhovas    \\
\foreignlanguage{ukrainian}{Чернівці}   & Chernivtsi   & Makiyivka      & Černivci   & Černovci      & Černivciai   & -             \\
\foreignlanguage{ukrainian}{Горлівка}   & Horlivka     & -              & Horlivka   & Gorlovka      & Horlivka     & -             \\
\foreignlanguage{ukrainian}{Кам'янське} & Kamianske    & -              & Kamjanska  & Kamjanske     & Kamianskė    & -             \\
\foreignlanguage{ukrainian}{Тернопіль}  & Ternopil     & Tarnopol       & Ternopiļa  & Ternopole     & Ternopilis   & Ternopolis   \\ \bottomrule \\
 \end{tabular}
 \end{table}
 
 \begin{table}[]
\centering
\caption{Examples of Ukrainian-Latvian and Ukrainian-Lithuanian toponym translation with and without terminology integration (incorrect translations of toponyms are underlined).}
 \label{table:example}
\begin{tabular}{rl} \toprule
\multicolumn{2}{c}{\textbf{Ukrainian-Latvian}}                                                                              \\ \midrule
\textbf{Source:} &
  \begin{tabular}[c]{@{}l@{}}\foreignlanguage{ukrainian}{Київ, Харків, Одеса, Дніпро і Донецьк мають пять}\\ \foreignlanguage{ukrainian}{найбільших українських міст.}\end{tabular} \\
\textbf{Without terms:} &
  \begin{tabular}[c]{@{}l@{}}\underline{Kijeva}, \underline{Harkova}, Odesa, \underline{Dņepra} un \underline{Doņecka} ir piecas lielākās\\ Ukrainas pilsētas.\end{tabular} \\
\textbf{With terms:} &
  \begin{tabular}[c]{@{}l@{}}Kijiva, Harkiva, Odesa, Dnipro un Donecka ir piecas lielākās\\ Ukrainas pilsētas.\end{tabular} \\ \midrule
\multicolumn{2}{c}{\textbf{Ukrainian-Lithuanian}}                                                                           \\ \midrule
\textbf{Source:} &
  \begin{tabular}[c]{@{}l@{}}\foreignlanguage{ukrainian}{Ми доставляємо товари в Запоріжжя, Львів, Чернігів та}\\ \foreignlanguage{ukrainian}{Тернопіль.}\end{tabular} \\
\textbf{Without terms:} & Mes pristatome prekes į \underline{Zaporožę}, \underline{Lvovą}, \underline{Černigą} ir Ternopilį. \\
\textbf{With terms:}    & Prekes pristatome į Zaporižę, \underline{Lvovą}, Černyhivą ir Ternopilį. \\           \bottomrule
\end{tabular}
\end{table}

\subsection{Ukrainian Toponym Translation}
Historically Ukrainian toponyms in the languages of the Baltic states have been introduced via Russian. Thus traditionally, Latvian and Lithuanian representations of Ukrainian toponyms have leaned on the conventions of Russian pronunciation. Traditions, however, are subject to cultural changes, as exemplified by the decommunization of Ukrainian toponymy after the collapse of the USSR and the proclamation of independent Ukraine \citep{demska2020urbanonimia}. Likewise, shifts in geopolitical allegiances can also be a decisive factor in changing language customs. Here, the example is the departure from the Russian-based representations of Ukrainian toponyms in Latvian to favour Ukrainian-based pronunciation. Since 2014 when the Russian Federation started military aggression against Ukraine, the expert committee of the Latvian State Language Centre has twice pushed for changes in the Latvian language representations of Ukrainian city names -- first in 2017\footnote{Accessed May 5, 2022 \url{https://www.vestnesis.lv/op/2017/208.18}} and then in 2019.\footnote{Accessed May 5, 2022 \url{https://www.vestnesis.lv/op/2019/42.37}} The final decision to officially stop using Russian-based representations of Ukrainian city names was reached on March 9, 2022,\footnote{Accessed May 5, 2022 \url{https://www.vestnesis.lv/op/2022/67.4}} only two weeks after the Russian invasion of Ukraine. 

Although these swift decisions reflect the political climate and the sentiment of the people, these changes hardly have had time to reach the training data of data-driven natural language processing tools. So we take advantage of our MT system's dynamic terminology integration capability and approach the problem of Ukrainian toponym translation as a terminology integration task. Specifically, we prepare toponym glossaries (see Table~\ref{table:toponym_translations}) mapping both the new and obsolete terms to their new and preferred translations. Before translating, we compare the stemmed version of each word in the sentence against the stemmed Ukrainian toponyms in the glossary and annotate them with their preferred translation if we find one. Then, we pass the annotated sentence to a system that is trained with TLA and can use the annotations to translate and inflect the toponym according to the sentence context. For more details, refer to \cite{bergmanis-pinnis-2021-facilitating} and \cite{bergmanis-pinnis-2021-dynamic}.

Terminology integration using TLA applies soft constraints on an NMT model. Contrary to methods that apply hard constraints, e.g., constrained decoding \citep{post2018fast}, this enables the NMT model to have flexibility in how the annotations are used. The NMT system can freely decide on the most suitable inflected form for the given morphosyntactic context. However, this also means that in some cases, the NMT model can choose to ignore the annotations if there is a stronger internal signal for a different lexical choice. Table~\ref{table:example} shows two examples where Ukrainian toponyms are translated from Ukrainian into Latvian and Lithuanian. The example shows that terminology integration improves toponym translation quality for most cases except for one example,`\foreignlanguage{ukrainian}{Львів}', was translated using the obsolete variant. Nevertheless, we believe that soft constraints are more appropriate for morphologically rich languages. There is room for future work to reduce cases where the NMT model decides not to rely on the annotations.


\section{Conclusions and Discussion}
We examined the quality, usage patterns and translation volume of six low-resource MT systems translating between the Ukrainian language and each of the official languages of the Baltic states. Although the translation quality analysis revealed that our systems are no better than the other publicly available alternatives, the MT usage statistics showed that the general public nevertheless uses some of our MT systems. Meanwhile, other MT systems are integrated into Lithuania's governmental websites or used by government translators in Latvia. 

We found that the different approaches to MT integration in public services have led to vastly different volumes of translation requests. In Lithuania, whenever a user accesses a certain page on a website, the MT system translates its content, generating many translation requests. This method provides flexible and instant multilingualism on a website regardless of which page a user wants to access. In Latvia, the systems are used mainly by translators and public service officials in MT post-editing scenarios. While this approach generates fewer translation requests, it also limits what content users can access in their native language. 

Knowing which systems are used most actively, we revisit them to improve their quality. Because of the different ways systems are integrated, in contrast to the Latvian-Ukrainian system,  the translations of the Lithuanian-Ukrainian system reach users without being checked by professional translators. Motivated by this finding, we retrained the Lithuanian-Ukrainian MT system using nearly twenty million sentences of back-translated data, which allowed us to, as measured by automatic metrics, outperform eTranslation and close the gap with Google Translate. Finally, we cast the Ukrainian toponym translation as a terminology integration task and show how to dynamically solve the changing and divergent spelling of place names when systems are deployed. 

There are options for future work to improve the MT between the official languages of Baltic states and Ukrainian beyond the quality achieved within this work. One, evident from Table~\ref{tab:data-stats}, is to obtain more high-quality data. Indeed, the amount of training data after filtering ranges from two to nearly five million parallel sentences, which is not much compared to other European language pairs. Another potential avenue for future work is to train multilingual MT models \citep{dabre2020survey} translating from many source languages to one target language. In such a setup, including one resource-rich language pair, such as English-Ukrainian, could help via means of transfer-learning \citep{kocmi2019exploring}, or at the very least, as a form of regularization \citep{neubig-hu-2018-rapid}. 

While this work provides a novel analysis of the MT usage in Baltic states to address language barriers rising from a refugee crisis, there are other similar efforts to use language technology to aid people displaced by the Russian war in Ukraine. One such effort is ÚFAL for Ukraine by Charles University in Prague, which offers an MT system for Czech-Ukrainian, Charles Translator for Ukraine. Their MT system builds on the previous work on Czech-English MT \citep{popel2020transforming} and can be accessed on the web,\footnote{Accessed July 14, 2022 \url{https://lindat.cz/translation}} via an android app,\footnote{Accessed July 14, 2022 \url{https://play.google.com/store/apps/details?id=cz.cuni.mff.ufal.translator}} as well as in the form of chatbots for Telegram and Messenger and other messaging services.\footnote{Accessed July 14, 2022 \url{https://github.com/martin-majlis/uk-cs-translation-bot}}

\section*{Acknowledgements} 

The research has been supported by the European Regional Development Fund within the research project ``AI Assistant for Multilingual Meeting Management'' No. 1.1.1.1/ 19/A/082.

\bibliography{references}

\end{document}